\documentclass[10pt,twocolumn,letterpaper]{article}
\usepackage{cvpr}
\usepackage{times}
\usepackage{epsfig}
\usepackage{helvet}
\usepackage{courier}
\usepackage{floatrow}
\usepackage{multirow}
\newfloatcommand{capbtabbox}{table}[][\FBwidth]
\usepackage{comment}
\usepackage[utf8]{inputenc}
\usepackage[T1]{fontenc}  
\usepackage{url}      
\usepackage{booktabs}     
\usepackage{amsfonts}    
\usepackage{units}      
\usepackage{microtype}
\usepackage{xcolor}
\usepackage{colortbl,booktabs}
\usepackage{times}
\usepackage{soul}
\usepackage{url}
\usepackage[small]{caption}
\usepackage{graphicx}
\usepackage{amsmath}
\usepackage{booktabs}
\usepackage{algorithm}
\usepackage{algorithmic}
\usepackage{epstopdf}
\usepackage{amsthm}
\usepackage{amssymb}
\usepackage{subfigure}
\usepackage{wrapfig}
\newcommand{\tabincell}[2]{\begin{tabular}{@{}#1@{}}#2\end{tabular}}

\usepackage[flushleft]{threeparttable}

\usepackage[hang,flushmargin]{footmisc}

\usepackage[pagebackref=true,breaklinks=true,letterpaper=true,colorlinks,bookmarks=false]{hyperref}

\cvprfinalcopy 

\def\cvprPaperID{6014} 

\ifcvprfinal\fi
\begin{document}

\title{Forward and Backward Information Retention\\for Accurate Binary Neural Networks}

\author{Haotong Qin\textsuperscript{1}, Ruihao Gong\textsuperscript{1}, Xianglong Liu\textsuperscript{1,2\thanks{Corresponding author}}, Mingzhu Shen\textsuperscript{1}, Ziran Wei\textsuperscript{4}, Fengwei Yu\textsuperscript{3}, Jingkuan Song\textsuperscript{5}\\
\textsuperscript{1}{State Key Lab of Software Development Environment, Beihang University}\\
\textsuperscript{2}{Beijing Advanced Innovation Center for Big Data-Based Precision Medicine, Beihang University}\\
\textsuperscript{3}{SenseTime Research}\quad
\textsuperscript{4}{Beijing University of Posts and Telecommunications}\\
\textsuperscript{5}{Center for Future Media, University of Electronic Science and Technology of China}\\
{\small \tt \{qinhaotong, gongruihao, xlliu\}@nlsde.buaa.edu.cn, {lavieenrosesmz@outlook.com},}\\
{\small\tt {yufengwei@sensetime.com}, \{weiziran125, jingkuan.song\}@gmail.com}
}

\maketitle

\begin{abstract}
\vspace{-0.1in}
Weight and activation binarization is an effective approach to deep neural network compression and can accelerate the inference by leveraging bitwise operations. Although many binarization methods have improved the accuracy of the model by minimizing the quantization error in forward propagation, there remains a noticeable performance gap between the binarized model and the full-precision one. Our empirical study indicates that the quantization brings information loss in both forward and backward propagation, which is the bottleneck of training accurate binary neural networks. To address these issues, we propose an \textbf{I}nformation \textbf{R}etention \textbf{Net}work (IR-Net) to retain the information that consists in the forward activations and backward gradients. IR-Net mainly relies on two technical contributions: (1) \textit{Libra Parameter Binarization} (Libra-PB): simultaneously minimizing both quantization error and information loss of parameters by balanced and standardized weights in forward propagation; (2) \textit{Error Decay Estimator} (EDE): minimizing the information loss of gradients by gradually approximating the $\mathtt{sign}$ function in backward propagation, jointly considering the updating ability and accurate gradients. We are the first to investigate both forward and backward processes of binary networks from the unified information perspective, which provides new insight into the mechanism of network binarization. Comprehensive experiments with various network structures on CIFAR-10 and ImageNet datasets manifest that the proposed IR-Net can consistently outperform state-of-the-art quantization methods. 
\vspace{-0.1in}
\end{abstract}

\section{Introduction}

Deep neural networks (DNNs), especially convolutional neural networks (CNNs), have been well demonstrated in a wide variety of computer vision applications such as image classification~\cite{krizhevsky2012imagenet,VeryDeepConvolutional,7298594,wang2019dynamic,Wang_2019_ICCV,Yang_2020_CVPR,Zhu_2020_CVPR,Wu_2020_CVPR}, object detection~\cite{DBLP:journals/corr/GirshickDDM13,DBLP:journals/corr/Girshick15,DBLP:journals/corr/abs-1904-02701,NIPS2015_5638,Li_2019_CVPR} and semantic segmentation~\cite{Everingham:2010:PVO:1747084.1747104,Zhuang_2019_CVPR}. Traditional CNNs are usually with massive parameters and high computational complexity for the requirement of high accuracy. Consequently, deploying the most advanced deep CNN models requires expensive storage and computing resources, which largely limits the applications of DNNs on portable devices such as mobile phones and cameras.
Binary neural networks are appealing to the community for their tiny storage usage and efficient inference~\cite{NIPS2015_5647,CI-BCNN,CBCN,BENN,Xu_2019_CVPR,Gu_2019_ICCV}, which results from the binarization of both weights and activations and the efficient convolution implemented by bitwise operations.
Although much progress has been made on binarizing DNNs, the existing quantization methods remains a significant drop of accuracy compared with the full-precision counterparts~\cite{Dong_2019_ICCV,Morozov_2019_ICCV,Ajanthan_2019_ICCV,Jung_2019_CVPR,Yang_2019_CVPR,Wang_2019_CVPR,Cao_2019_CVPR,Nagel_2019_ICCV,Qin_2020_pr}.

The performance degradation of binary neural networks is mainly caused by the limited representation ability and discreteness of binarization, which results in severe information loss in both forward and backward propagation. In the forward propagation, when the activations and weights are restricted to two values, the model's diversity sharply decreases, while the diversity is proved to be the key of pursuing high accuracy of neural networks~\cite{diverse}.
Two approaches are widely used to increase the diversity of neural networks: increasing the number of neurons or increasing the diversity of feature maps.
For example, Bi-Real Net~\cite{Liu_2018_ECCV} is targeted at the latter by adding on a full-precision shortcut to the quantized activations, which achieves significant performance improvement.
However, with the additional floating-point add operation, Bi-Real Net inevitably faces worse efficiency than vanilla binary neural networks.

The diversity implies the ability to carry enough information during the forward propagation, meanwhile, accurate gradients in the backward propagation provide correct information for optimization. However, during the training process of binary neural networks, the discrete binarization always leads to inaccurate gradients and the wrong optimization direction. 
To better deal with the discreteness, different approximations of binarization for backward propagation have been studied~\cite{DBLP:conf/cvpr/CaiHSV17,Liu_2018_ECCV,BNN+,selfBN,ImprovedTraining}, mainly categorized into either improving the updating ability or reducing the mismatching area between $\mathtt{sign}$ function and the approximation one.
Unfortunately, the difference between the early and later training stages is always be ignored, where in practice strong updating ability is usually highly required when the training process starts and small gradient error becomes more important at the end of the training. 
It is insufficient to acquire as much information reflected from the loss function as possible only focusing on one point.

\begin{figure}[tp!]
    \vspace{-0.11in}
	\begin{center}
		\includegraphics[width=0.9\linewidth]{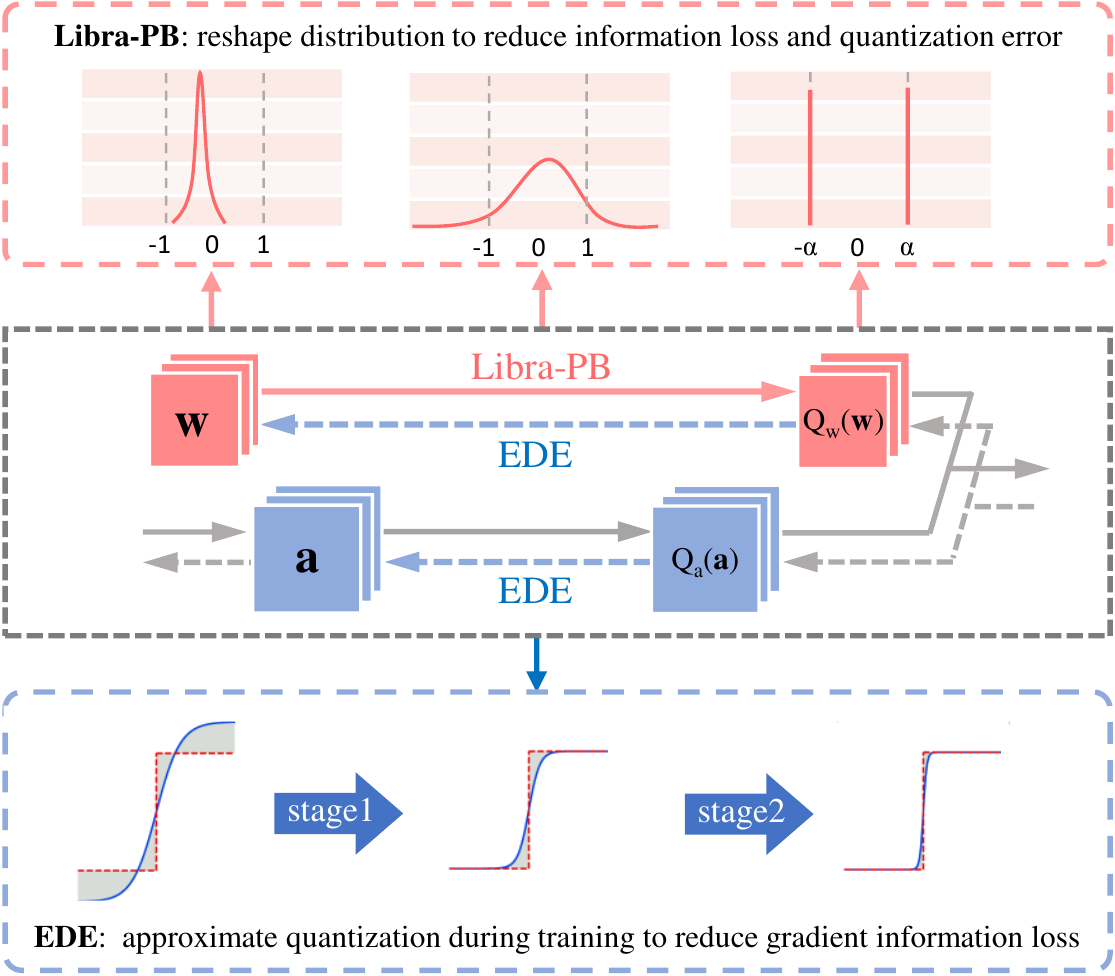}
	\end{center}
	\vspace{-0.2in}
	\caption{Overview of our IR-Net training for a convolutional layer,
	consisting of Libra Parameter Binarization (Libra-PB) in forward propagation and Error Decay Estimator (EDE) in backward propagation.
	Libra-PB changes the weight distribution in forward propagation to retain the information of weights and activations. And the shape change of EDE during the entire training process reduces the gradient information loss in backward propagation.}
	\label{fig:data_flow}
	\vspace{-0.05in}
\end{figure}

To solve the above-mentioned problems, this paper is the first to study the model binarization from the view of information flow and proposes a novel \textbf{I}nformation \textbf{R}etention \textbf{Net}work (IR-Net) (see the overview in Fig~\ref{fig:data_flow}). Our goal is to train highly accurate binarized models by retaining the information in the forward and backward propagation: (1) IR-Net introduces a balanced and standardized quantization method called \textit{Libra Parameter Binarization} (Libra-PB) in the forward propagation. With Libra-PB, we can minimize the information loss in forward propagation by maximizing the information entropy of the quantized parameters and minimizing the quantization error, which ensures a high diversity. (2) In backward propagation, IR-Net adopts the \textit{Error Decay Estimator} (EDE) to calculate gradients and minimizes the information loss by better approximating the $\mathtt{sign}$ function, which ensures sufficient updating at the beginning and accurate gradients at the end of the training.

Our IR-Net presents a new and practical perspective to understand how the binarized network works. Besides the strong capability of preserving the information forward/backward in the deep network, it also enjoys good versatility and can be optimized in a standard network training pipeline. We evaluate our IR-Net with image classification tasks on the CIFAR-10 and ImageNet datasets. The experimental results show that our method performs remarkably well across various network structures such as ResNet-20, VGG-Small, ResNet-18, and ResNet-34, surpassing previous quantization methods by a wide margin. Our code is released at \href{https://github.com/htqin/IR-Net}{GitHub}.

\section{Related Work}

Network binarization aims to accelerate the inference of neural networks and save memory occupancy without much accuracy degradation. One approach to speed up low-precision networks is to utilize bitwise operations. 
By directly binarizing the 32-bit parameters in DNNs including weights and activations, we can achieve significant accelerations and memory reductions.
XNOR-Net~\cite{DBLP:conf/eccv/RastegariORF16} utilizes a deterministic binarization scheme and minimizes the quantization error of the output matrix by employing some scalars in each layer. TWN~\cite{DBLP:journals/corr/LiL16} and TTQ~\cite{DBLP:journals/corr/ZhuHMD16} enhance the representation ability of neural networks with more available quantization points. 
ABC-Net~\cite{ABCNet} recommends using more binary bases for weights and activations to improve accuracy, while compression and acceleration ratios are reduced accordingly. \cite{DBLP:conf/cvpr/CaiHSV17} proposed HWGQ considering the quantization error from the aspect of activation function. \cite{LQ-Net} further proposed LQ-Net with more training parameters, which achieved comparable results on the ImageNet benchmark but increased the memory overhead.

Compared with other model compression methods, \eg, pruning~\cite{han2015learning,han2016deep,he2017channel} and matrix decomposition~\cite{yu2017on,DBLP:journals/pami/WangXXT19}, network binarization can greatly reduce the memory consumption of the model, and make the model fully compatible with bitwise operations to get good acceleration.
Although much progress has been made on network binarization, the existing quantization methods still cause a significant drop of accuracy compared with the full-precision models, since great information loss still exists in the training of binary neural networks. Therefore, to retain the information and ensure a correct information flow during the forward and backward propagation of binarized training, IR-Net is designed.

\section{Preliminaries}

The main operation in deep neural networks is expressed as:
\begin{equation}
z=\mathbf w^{\top}\mathbf a ,
\end{equation}
where $\mathbf w \in \mathbb R^n$ indicates the weight vector, $\mathbf a\in\mathbb R^n$ indicates the input activation vector computed by the previous network layer.

The goal of network binarization is to represent the floating-point weights and/or activations with 1-bit. In general, the quantization can be formulated as:
\begin{equation}
\label{eq:quantized}
Q_x(\mathbf x)=\alpha\mathbf{B_x} ,
\end{equation}
where $\mathbf x$ denotes floating-point parameters including floating-point weights $\mathbf w$ and activations $\mathbf a$, and $\mathbf B_{\mathbf x}\in{\{-1, +1\}}^n$ denotes binary values including binary weights $\mathbf B_{\mathbf w}$ and activations $\mathbf B_{\mathbf a}$. $\alpha$ denotes scalars for binary values including ${\alpha}_{\text{w}}$ for weights and ${\alpha}_{\text{a}}$ for activations. And we usually use $\mathtt{sign}$ function to get $\mathbf B_{\mathbf x}$ :
\begin{equation}
\label{eq:binarized}
\mathbf{B_x}=\mathtt{sign}(\mathbf{x})=
\begin{cases}
+1,& if \ \mathbf x \ge 0\\
-1,& otherwise.
\end{cases}
\end{equation}

With the quantized weights and activations, the vector multiplications in the forward propagation can be reformulated as
\begin{equation}
z = Q_w(\mathbf w)^\top Q_a(\mathbf a)= {{\alpha}_{\text{w}}}{{\alpha}_{\text{a}}}({\mathbf{B_w}}\odot {\mathbf{B_a}}),
\end{equation} 
where $\odot$ denotes the inner product for vectors with bitwise operations XNOR and Bitcount.

In the backward propagation, the derivative of the $\mathtt{sign}$ function is zero almost everywhere, which makes it incompatible with backward propagation, since exact gradients for the original values before the discretization (pre-activations or weights) would be zeroed.
So "Straight-Through Estimator (STE) \cite{bengio2013estimating}" is generally used to train binary models, which propagates the gradient through $\mathtt{Identity}$ or $\mathtt{Hardtanh}$ function.

\section{Information Retention Network}
In the paper, we point out that the bottleneck of training highly accurate binary neural networks mainly lies in the severe information loss of the training process. Information loss caused by the forward $\mathtt{sign}$ function and the backward approximation for gradient greatly harms the accuracy of binary neural networks. In this paper, we propose a novel model, \textbf{I}nformation \textbf{R}etention \textbf{Net}work (IR-Net), which retains the information in the training process and acquires highly accurate binarized models.

\subsection{Libra Parameter Binarization in Forward Propagation} 

In the forward propagation, the quantization operation brings information loss. Many quantized convolutional neural networks, including binarized models~\cite{DBLP:conf/eccv/RastegariORF16,DBLP:journals/corr/abs-1708-08687,LQ-Net}, find the optimal quantizer by minimizing the quantization error :
\begin{equation}\label{quantizatoin_error}
\min J(Q_x(\mathbf x))=\lVert{\mathbf x-Q_x(\mathbf x)}\rVert^2,
\end{equation}
where $\mathbf x$ indicates the full-precision parameters, $Q_x{(\mathbf x)}$ denotes the quantized parameters and $J(Q_x(\mathbf x))$ denotes the quantization error between full-precision and binary parameters. The objective function (Eq.~(\ref{quantizatoin_error})) assumes that quantized models should completely follow the pattern of full-precision models. However, this is not always true especially when extremely low bit-width is applied. For binary models, the representation ability of their parameters is limited to two values, which makes the information carried by neurons easy to lose. The solution space of binary neural networks also quite differs from that of full-precision neural networks. Therefore, without retaining the information through networks, it is insufficient and difficult to promise a good binarized network only by minimizing the quantization error.

To retain the information and minimize the information loss in forward propagation, we propose Libra Parameter Binarization (Libra-PB) that jointly considers both quantization error and information loss.
For a random variable $b\in\{-1, +1\}$ obeying Bernoulli distribution, whose probability mass function is
\begin{equation}
f(b)=
\begin{cases}
p , \quad &if\ b = +1\\
1-p , \quad &if\ b = -1,
\end{cases}
\end{equation}
where $p$ is the probability of taking the value +1, $p\in(-1,1)$, and each element in $\mathbf{B_x}$ can be viewed as a sample of $b$. The entropy of $Q_x(\mathbf x)$ in Eq.~(\ref{eq:quantized}) can be calculated by:
\begin{equation}\label{ie}
\mathcal{H}(Q_x(\mathbf{x}))=\mathcal{H}(\mathbf{B_x})=-p\ln(p)-(1-p)\ln(1-p).
\end{equation}
If we only pursue the goal of minimizing quantization error, the information entropy of the quantized parameters can be close to zero in extreme case. Therefore, Libra-PB combines the quantization error and information entropy of quantized values as objective function, which is defined as
\begin{equation}\label{mil_f}
\min J(Q_x(\mathbf x))-\lambda \mathcal{H}(Q_x(\mathbf x)).
\end{equation}
Under the Bernoulli distribution assumption, when $p = 0.5$, the information entropy of the quantized values takes the maximum value. This means the quantized values should be evenly distributed. Therefore, we balance weights with zero-mean attribute by subtracting the mean of full-precision weights. Moreover, to make the training more stable without negative effect deriving from the weight magnitude and thus the gradient, we further normalize the balanced weight. The standardized balanced weights $\hat{\mathbf w}_{\text{std}}$ are obtained through standardization and balance operations as follows:
\begin{equation}
\hat{\mathbf w}_{\text{std}}=\frac{\hat{\mathbf w}}{\sigma(\hat{\mathbf w})},\quad
\hat{\mathbf w}=\mathbf w-\overline{\mathbf w}.
\end{equation}
where the $\sigma(\cdot)$ denotes the standard deviation. $\hat{\mathbf w}_{\text{std}}$ has two characteristics: (1) $zero\verb|-|mean$, which maximizes the obtained binary weights' information entropy. (2) $unit\verb|-|norm$, which makes the full-precision weights involved in binarization more dispersed. Therefore, compared with the direct use of the balanced progress, the use of standardized balanced progress makes the weights steadily updated, and makes the binary weights $Q_w(\hat{\mathbf w}_{\text{std}})$ more stable during the training.
 
Since the value of $Q_w(\hat{\mathbf w}_{\text{std}})$ depends on the sign of $\hat{\mathbf w}_{\text{std}}$ and the distribution of $\mathbf{w}$ is almost symmetric~\cite{Simultaneously-Optimizing-Weight,ACIQ}, 
the balanced operation can maximize the information entropy of quantized $Q_w(\hat{\mathbf w}_{\text{std}})$ on the whole.
And when Libra-PB is used for weights, the information flow of activations in the network can also be maintained. Supposing quantized activations $Q_a(\mathbf a)$ have mean $\mathbb{E}[Q_a(\mathbf a)] = \mu\mathbf 1$, the mean of $z$ can be calculated by:
\begin{equation}
\mathbb{E}[z] = {Q_w(\hat{\mathbf w}_{\text{std}})}^\top \mathbb{E}[Q_a(\mathbf a)] = {Q_w(\hat{\mathbf w}_{\text{std}})}^\top\mu\mathbf 1.
\end{equation}
Because of using Libra-PB for weights in each layer, we have ${Q_w(\hat{\mathbf w}_{\text{std}})}^\top \mathbf 1 = 0$, the mean of output is zero. Therefore, the information entropy of activations in each layer can be maximized, which means that the information in activations can be retained.

To further minimize the quantization error and avoid extra expensive floating-point calculation in previous binarization methods, Libra-PB introduces an integer bit-shift scalar $s$ to expand the representation ability of binary weights. The optimal bit-shift scalar can be solved by:
\begin{equation}
\mathbf{B_{w}^*},s^*=\mathop{\arg\!\min}\limits_{\mathbf{B_{w}}, s}\lVert{\hat{\mathbf w}_{\text{std}}-\mathbf{B_{w}}\ll\gg s}\rVert^2\quad s.t.\  s\in\mathbb N
\end{equation}
where $\ll\gg$ stands for left or right bit-shift. $\mathbf{B_{w}^*}$ is calculated by $\mathbf{B_{w}^*}=\mathtt{sign}(\mathbf{\hat w}_{\text{std}})$, thus $s^*$ can be solved as:
\begin{equation}
    s^*=\mathtt{round}(\log_2 ({{\|\hat{\mathbf w}_{\text{std}}\|_{1}}/{n}})).
\end{equation}
where $n$ and $\|\hat{\mathbf w}_{\text{std}}\|_{1}$ denote the dimension and L1-norm of the vector, respectively.

\begin{figure}[tbp]
\vspace{-0.1in}
\includegraphics[width=0.9\textwidth]{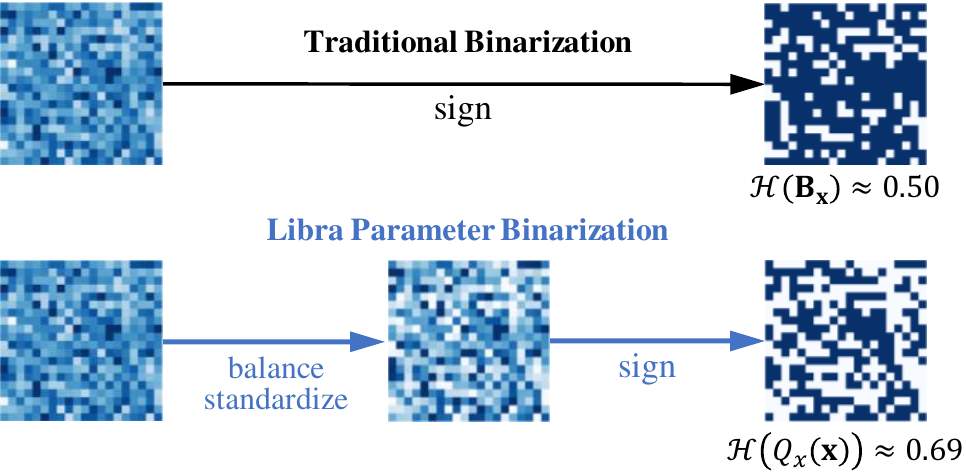}
\vspace{-0.15in}
\caption{Comparison on information entropy of binary weights quantized with Libra-PB $Q_x(\mathbf x)$ and $\mathtt{sign}$ function, respectively. Owing to the balance characteristic brought by Libra-PB, the information entropy of $Q_x(\mathbf x)$ is larger than $\mathtt{sign}(\mathbf{x})$, where $Q_x(\mathbf x)$ and $\mathtt{sign}(\mathbf{x})$ have a probability of 0.5 and 0.2 to take value 1 under Bernoulli distribution, respectively.}
\label{fig:Libra-PB}
\end{figure}

Therefore, our Libra Parameter Binarization for the forward propagation can be presented as below:
\begin{equation}
\label{eq:Libra-PB}
\begin{aligned}
Q_w(\hat{\mathbf w}_{\text{std}})=\mathbf{B_{w}}{\ll\gg}{s}&=\mathtt{sign}(\hat{\mathbf w}_{\text{std}}){\ll\gg}{s},\\
Q_a(\mathbf a)=\mathbf{B_{a}}&=\mathtt{sign}(\mathbf a).
\end{aligned}
\end{equation}
The main operations in IR-Net can be expressed as:
\begin{equation}
z=({\mathbf{B_{w}}}\odot {\mathbf{B_{a}}})\ll\gg s.
\end{equation}

As shown in Fig~\ref{fig:Libra-PB}, the parameters quantized by Libra-PB have the maximum information entropy under the Bernoulli distribution. We call our binarization method "Libra Parameter Binarization" because the parameters are balanced before the binarization to retain information.

Note that Libra-PB serves an implicit rectifier that reshapes the data distribution before binarization. In the literature, a few studies also realized this positive effect on the performance of BNNs and adopted empirical settings to redistribute parameters~\cite{DBLP:conf/eccv/RastegariORF16,Regularize-act-distribution}. For example, \cite{Regularize-act-distribution} proposed the specific degeneration problem of binarization and solved it using a specially designed additional regularization loss. Different from these works, we first straightforwardly present the information view to rethink the impact of parameter distribution before binarization, and promise the optimal solution by maximizing the information entropy. Moreover, in this framework, Libra-PB can accomplish the distribution adjustment by simply balancing and standardizing the weights before the binarization. This means that our method can be easily and widely applied to various neural network architectures and be directly plugged into the standard training pipeline with a very limited extra computation cost.

\begin{figure*}[htbp]
\vspace{-0.4in}
\centering 
\subfigure[$\mathtt{Identity}$]{
\includegraphics[width=0.15\linewidth]{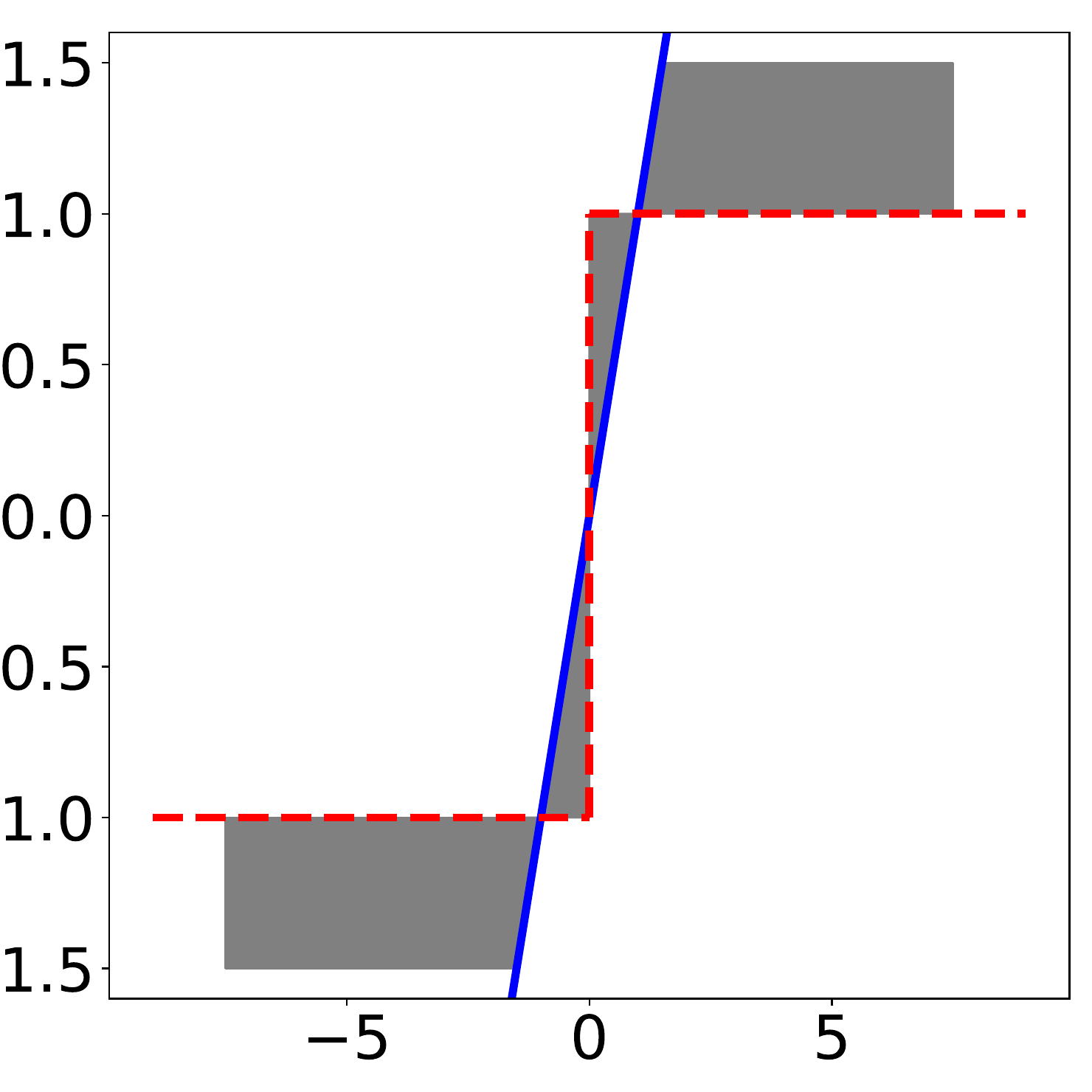}
}\hspace{0.1in}
\subfigure[$\mathtt{Clip}$]{
\includegraphics[width=0.15\linewidth]{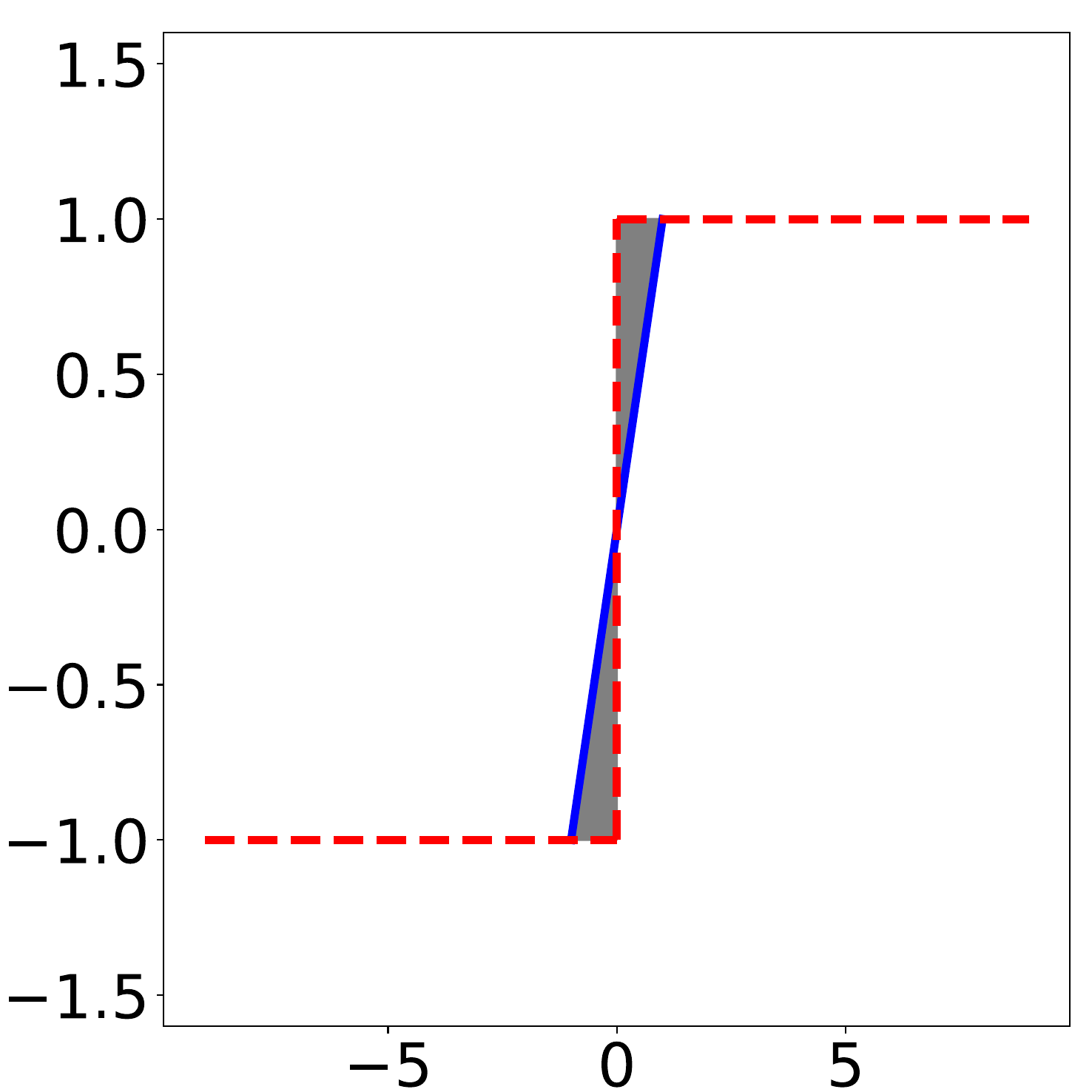}
}\hspace{0.1in}
\subfigure[EDE]{
\includegraphics[width=0.55\linewidth]{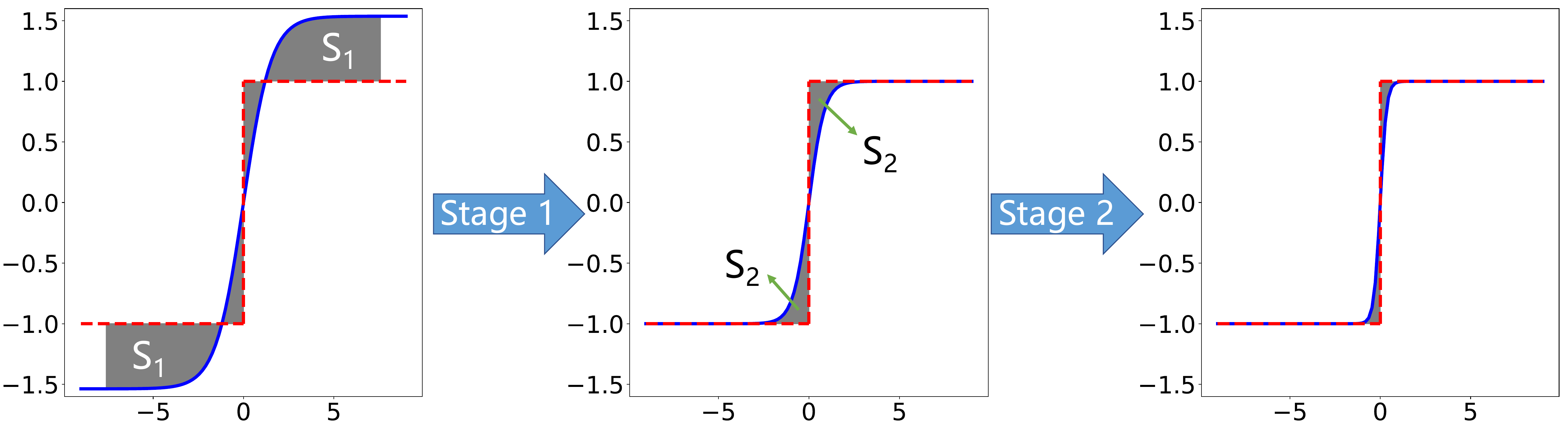}
}
\caption{Error caused by gradient approximation, represented by the area of gray shades. As is shown, (a) $\mathtt{Identity}$ approximation suffers huge error. (b) $\mathtt{Clip}$ approximation does not update the values outside the clipping interval. (c) EDE maintains the updating ability at early stage and progressively reduces the error. $S_1$ shrinks during Stage 1 by decreasing the clipping value and $S_2$ shrinks during Stage 2 by increasing the derivate.}
\label{fig:evolution}
\vspace{-0.2in}
\end{figure*}

\subsection{Error Decay Estimator in Backward Propagation}

Limited by the discontinuity of binarization, approximation of gradients is inevitable for the backward propagation. Thus, huge information loss is caused because the influence of quantization cannot be exactly modeled with the approximation. The approximation can be formulated as:
\begin{equation}
\frac{\partial \mathcal L}{\partial\mathbf w}=\frac{\partial \mathcal L}{\partial Q_w(\hat{\mathbf w}_{\text{std}})}\ \frac{\partial Q_w(\hat{\mathbf w}_{\text{std}})}{\partial \mathbf{w}} \approx \frac{\partial \mathcal L}{\partial Q_w(\hat{\mathbf w}_{\text{std}})}\ g'(\mathbf w),
\end{equation}
where $L(\mathbf w)$ represents the loss function, $g(\mathbf w)$ denotes the approximation of the $\mathtt{sign}$ function and $g'(\mathbf w)$ is the derivative of $g(\mathbf w)$. There are two common practices for approximation used in previous works:
\begin{equation}
    \mathtt{Identity}: y=x \quad \text{or} \quad
    \mathtt{Clip}: y=\mathtt{Hardtanh}(x).
\end{equation}  

The $\mathtt{Identity}$ function straightly passes the gradient information of output values to input values and completely neglects the effect of binarization. As shown in the shaded area of Fig~\ref{fig:evolution}(a), the gradient error is huge and will accumulate during the backward propagation. It is highly required to retain correct gradient information to avoid unstable training rather than ignore the noise caused by $\mathtt{Identity}$ for utilizing the Stochastic Gradient Descent Algorithm.

The $\mathtt{Clip}$ function takes the clipping attribute of binarization into account to reduce the gradient error. But it can only pass the gradient information inside the clipping interval. It can be seen in Fig~\ref{fig:evolution}(b) that for parameters outside [-1, +1], the gradient is clamped to 0. This means once the value jumps outside the clamping interval, it cannot be updated anymore. This characteristic greatly harms the updating ability of backward propagation, which can be proved by the fact that ReLU is a superior activation function compared with Tanh. Thus the $\mathtt{Clip}$ approximation increases the difficulty for optimization and decreases the accuracy in practice. It is crucial to ensure enough updating possibility, especially during the beginning of the training process.

$\mathtt{Identity}$ function loses the gradient information of quantization while $\mathtt{Clip}$ function loses the gradient information outside the clipping interval. There is a contradiction between these two kinds of gradient information loss. To make a balance and acquire the optimal approximation of backward gradient, we design Error Decay Estimator:
\begin{equation}
    g(x) = k\tanh{tx}
\end{equation}
where $g(x)$ is the backward approximation substitute for the forward $\mathtt{sign}$ function, $k$ and $t$ are control variables varying during the training process:
\begin{equation}
\centering
\label{eq:EDE}
\begin{aligned}
t=T_{\min}10^{\frac{i}{N}\times\log \frac{T_{\max}}{T_{\min}}},\quad k = \max(\frac{1}{t}, 1)
\end{aligned}
\end{equation}
where $i$ is the current epoch and $N$ is the number of epochs, $T_{\min}=10^{-1}$ and $T_{\max}=10^1$.

To retain the information deriving from loss function in the backward propagation, EDE introduces a progressive two-stage approach to approximate gradients. 

\noindent\textbf{Stage 1: Retain the updating ability of the backward propagation algorithm.} We keep the gradient estimation function's derivative value close to one, and then progressively reduce the clipping value from a large number to one. With this rule, our estimation function evolves from $\mathtt{Identity}$ to $\mathtt{Clip}$ approximation, which ensures the update ability at the early stage of training.

\noindent\textbf{Stage 2: Retain the accurate gradients for parameters around zero.} We keep the clipping value as one and gradually push the derivative curse to the shape of the staircase function. With this rule, our estimation function evolves from $\mathtt{Clip}$ approximation to $\mathtt{sign}$ function, which ensures the consistency of forward and backward propagation.

The shape change of EDE for each stage is shown in Fig~\ref{fig:evolution}(c). 
Our EDE updates all parameters in the first stage, and further makes the parameters more accurate in the second stage. Based on the two-stage estimation, EDE reduces the gap between the forward binarization function and the backward approximation function and meanwhile all parameters can be reasonably updated.

\begin{algorithm}[htbp]
	\caption{Forward and backward propagation for BNN training by the proposed IR-Net.} 
	\label{algo_forward}
	\small
	\begin{algorithmic}[1]
		\STATE \textbf{Require}: the input data $\mathbf a \in\mathbb R^{n}$, pre-activation $z \in\mathbb R$, full-precision weights $\mathbf w \in\mathbb R^{n}$.
		\STATE {\textbf{Forward propagation}}
		\STATE \quad Compute binary weight by Libra-PB [Eq.~(\ref{eq:Libra-PB})]:\\
		\quad \quad $\hat{\mathbf w}_{\text{std}}=\frac{\mathbf w-\overline{\mathbf w}}{\sigma(\mathbf w-\overline{\mathbf w}))}$,\quad $s^*=\mathtt{round}(\log_2 {\frac{\|\hat{\mathbf w}_{\text{std}}\|_{1}}{n}})$\\
		\quad \quad $Q_w(\hat{\mathbf w}_{\text{std}})=\mathbf{B_{w}}\ll\gg s =\mathtt{sign}(\hat{\mathbf w}_{\text{std}})\ll\gg s$\\
		\STATE \quad Compute balanced binary input data [Eq.~(\ref{eq:Libra-PB})]:\\
		\quad \quad $Q_a(\mathbf a)=\mathbf{B_{a}}=\mathtt{sign}(\mathbf a)$;
		\STATE \quad Calculate the output: ${z}=(\mathbf{B_{w}}\odot \mathbf{B_{a}})\ll\gg s$
		\STATE {\textbf{Back propagation}}
		\STATE \quad Update the $g'(\cdot)$ via EDE:\\
		\quad \quad Get current $t$ and $k$ by Eq.~(\ref{eq:EDE})\\
		\quad \quad Update the $g'(\cdot)$:\quad $g'(x)=kt(1-\tanh^2(tx))$
		\STATE \quad Calculate the gradients \wrt $\mathbf a$:\\
		\quad \quad $\frac{\partial{\mathcal{L}}}{\partial{\mathbf a}}=\frac{\partial{\mathcal{L}}}{\partial Q_a(\mathbf a)}g'({\mathbf a})$\\
		\STATE \quad Calculate the gradients \wrt $\mathbf w$:\\
		\quad \quad $\frac{\partial{\mathcal{L}}}{\partial{\mathbf w}}=\frac{\partial{\mathcal{L}}}{\partial Q_w(\hat{\mathbf w}_{\text{std}})}g'(\hat{\mathbf w}_{\text{std}})2^{s}$\\
		\STATE {\textbf{Parameters Update}}
		\STATE \quad Update $\mathbf w$: $\mathbf w=\mathbf w-\eta\frac{\partial{\mathcal{L}}}{\partial{\mathbf w}}$, where $\eta$ is learning rate.
	\end{algorithmic}
\end{algorithm}

\subsection{Analysis and Discussions}

The training process of our IR-Net is summarized in Algorithm~\ref{algo_forward}. In this section, we will analyze IR-Net from different aspects.

\begin{table}[htb]
	\caption{The additional floating-point operations consumed by different binarization methods.}
	\vspace{-0.14in}
	\centering
	\small
	\setlength{\tabcolsep}{3.8mm}{
	\begin{threeparttable}
	\begin{tabular}{lcc}
	    \toprule
		Method & Float Operations & \tabincell{c}{Bitwise Operations}\\
		\midrule
		XNOR-Net& $C_{\text{1}}$&$C_{\text{1}}\times C_{\text{2}}$\\ 
		LQ-Net&$C_{\text{1}}$&$C_{\text{1}}\times C_{\text{2}}$\\ 
		Ours&0&$C_{\text{1}}\times C_{\text{2}}+C_{\text{1}}$\\
	    \bottomrule
	\end{tabular}
	\begin{tablenotes}
    \footnotesize
    \item[*] $C_{\text{1}}=w_{\text{out}}\times h_{\text{out}}\times c_{\text{out}}$ and  $C_{\text{2}}=w_{\text{k}}\times h_{\text{k}}\times c_{\text{in}}$, where $c_{\text{out}} $, $ c_{\text{in}} $, $ w_{\text{k}} $, $ h_{\text{k}} $, $ w_{\text{out}} $, $ h_{\text{out}} $ denote the number of output channels, input channels, kernel width, kernel height, output width, and output height, respectively. The Bitwise operation mainly consists of XNOR, Bitcount and Bit-shift.
    \end{tablenotes}
    \end{threeparttable}
	}
	\label{tab:flops}
	\vspace{-0.2in}
\end{table}

\subsubsection{Complexity Analysis}
Since Libra-PB is applied on weights, there is extra operation for binarizing activations in IR-Net. And in Libra-PB, with the novel bit-shift scales, the computation costs are reduced compared with the existing solutions with floating-point scalars (\eg, XNOR-Net, and LQ-Net), as shown in Table~\ref{tab:flops}. Later, we further test the real speed of deployment on hardware and the results can be seen in the Deployment Efficiency Section.

\subsubsection{Stabilize Training}
\begin{figure}[tp!]
\vspace{-0.35in}
\centering
\includegraphics[width=0.7\linewidth]{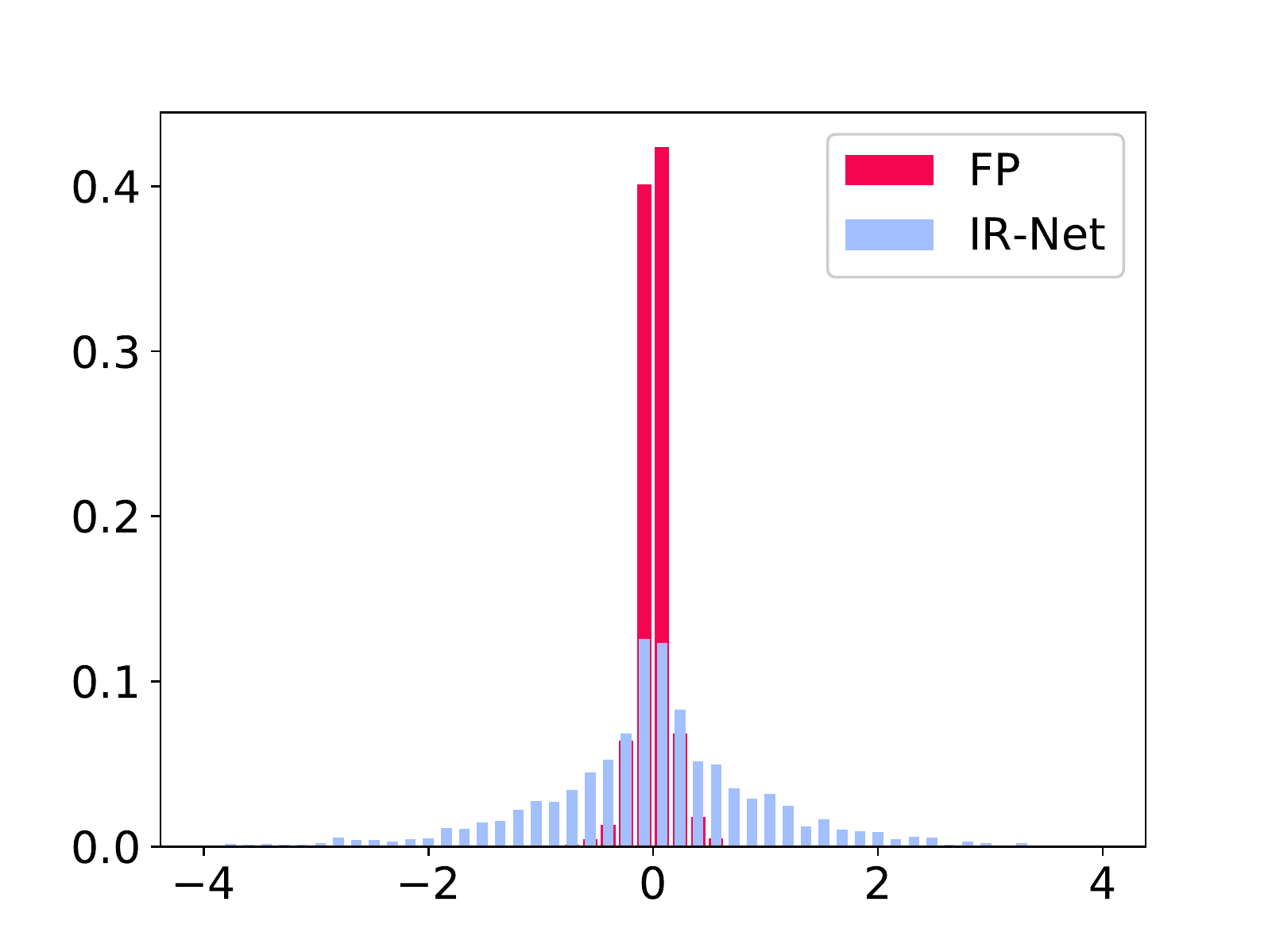}
\caption{Full-precision weights (in red) in neural networks have a small data range and always gather around 0, and thus their signs are highly possible to flip in backward propagation. IR-Net balances and standardizes the weights (in blue) before the binarization for stabilizing training.}
\label{distribution}
\vspace{-0.2in}
\end{figure}

In Libra Parameter Binarization, weight standardization is introduced for reducing the gap between full-precision weights and the binarized ones, avoiding the noise caused by binarization. Fig~\ref{distribution} shows the data distribution of weights without standardization, obviously more concentrated around 0. This phenomenon means the signs of most weights are easy to change during the process of optimization, which directly causes unstable training of binary neural networks. By redistributing the data, weight standardization implicitly sets up a bridge between the forward Libra-PB and backward EDE, contributing to a more stable training of binary neural networks.

\section{Experiments}

In this section, we conduct experiments on two benchmark datasets: CIFAR-10~\cite{CIFAR} and ImageNet (ILSVRC12)~\cite{Deng2009ImageNet} to verify the effectiveness of the proposed IR-Net and compare it with other state-of-the-art (SOTA) methods.

\textbf{IR-Net:} We implement our IR-Net using PyTorch because of its high flexibility and powerful automatic differentiation mechanism. When constructing a binarized model, we simply replace the convolutional layers in the origin models with the binary convolutional layer binarized by our method.

\textbf{Network Structures:} We employ the widely-used network structures including VGG-Small~\cite{LQ-Net}, ResNet-18, ResNet-20 for CIFAR-10, and ResNet-18, ResNet-34~\cite{he2016deep} for ImageNet. To prove the versatility of our IR-Net, we evaluate it on both the normal structure and the Bi-Real~\cite{Liu_2018_ECCV} structure of ResNet.
All convolutional and fully-connected layers except the first and last one are binarized, and we select $\mathtt{Hardtanh}$ as our activation function instead of ReLU when we binarize the activation.

\textbf{Initialization:} Our IR-Net is trained from scratch (random initialization) without leveraging any pre-trained model. To evaluate our IR-Net on various network architectures, we mostly follow the hyper-parameter settings of their original papers~\cite{DBLP:conf/eccv/RastegariORF16,LQ-Net,Liu_2018_ECCV}. Among the experiments, we apply SGD as our optimization algorithm.

\subsection{Ablation Study}
In this part, we investigate the behaviors and effects of the proposed Libra-PB and EDE techniques on BNN performance.

\subsubsection{Effect of Libra-PB}
\begin{figure}[tp!]
\vspace{-0.2in}
\includegraphics[width=0.7\textwidth]{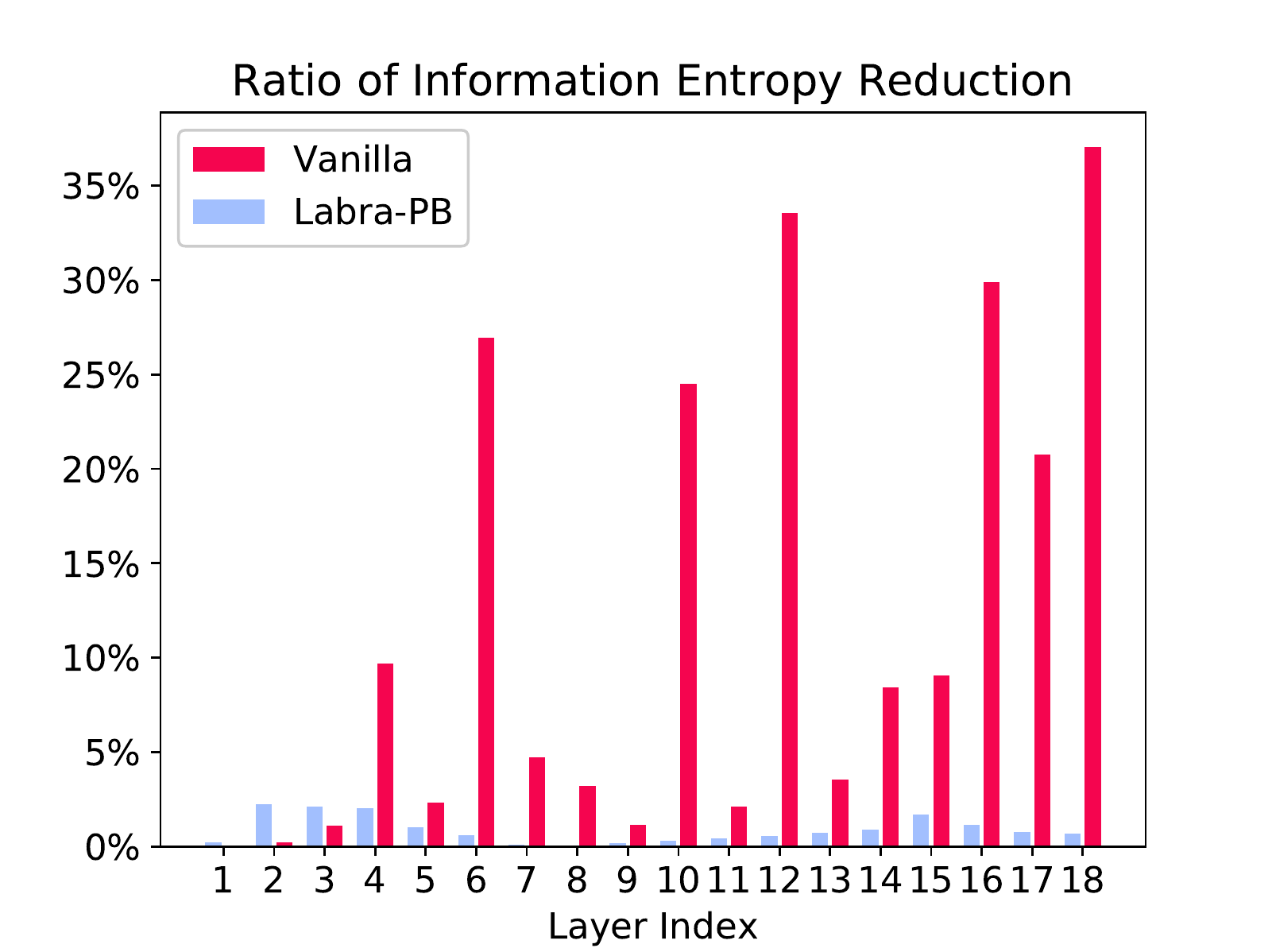}
\caption{Libra-PB's effect on information entropy of activations of each layer in ResNet-20. The ratio of information entropy reduction (compared with the maximum information entropy under the Bernoulli distribution) quantized by Libra-PB is significantly larger than vanilla binarization.}
\vspace{-0.1in}
\label{fig:activation_entropy}
\end{figure}

Our Libra-PB can maximize the information entropy of binary weights and binary activations in IR-Net by adjusting the distribution of weights in the network. Since an explicit balance operation is used before the binarization, the binary weights of each layer in the network have maximum information entropy. Binary activations affected by binary weights in IR-Nets also have maximum information entropy.
To demonstrate the information retention of Libra-PB in IR-Net, in Fig~\ref{fig:activation_entropy}, we show the information loss of each layer's binary activations in the networks quantized by vanilla binarization and Libra-PB. Vanilla binarization suffers a large reduction in the information entropy of the binary activations. In the network quantized by Libra-PB, the activation of each layer is close to the maximum information entropy under the Bernoulli distribution. And in the forward propagation, the information loss of binary activations caused by vanilla binarization accumulates layer by layer. Fortunately, the results in Fig~\ref{fig:activation_entropy} show that Libra-PB can retain the information in the binary activation of each layer.

\subsubsection{Effect of EDE}

\begin{figure}[tp!]
\vspace{-0.1in}
	\begin{center}
		\includegraphics[width=1\linewidth]{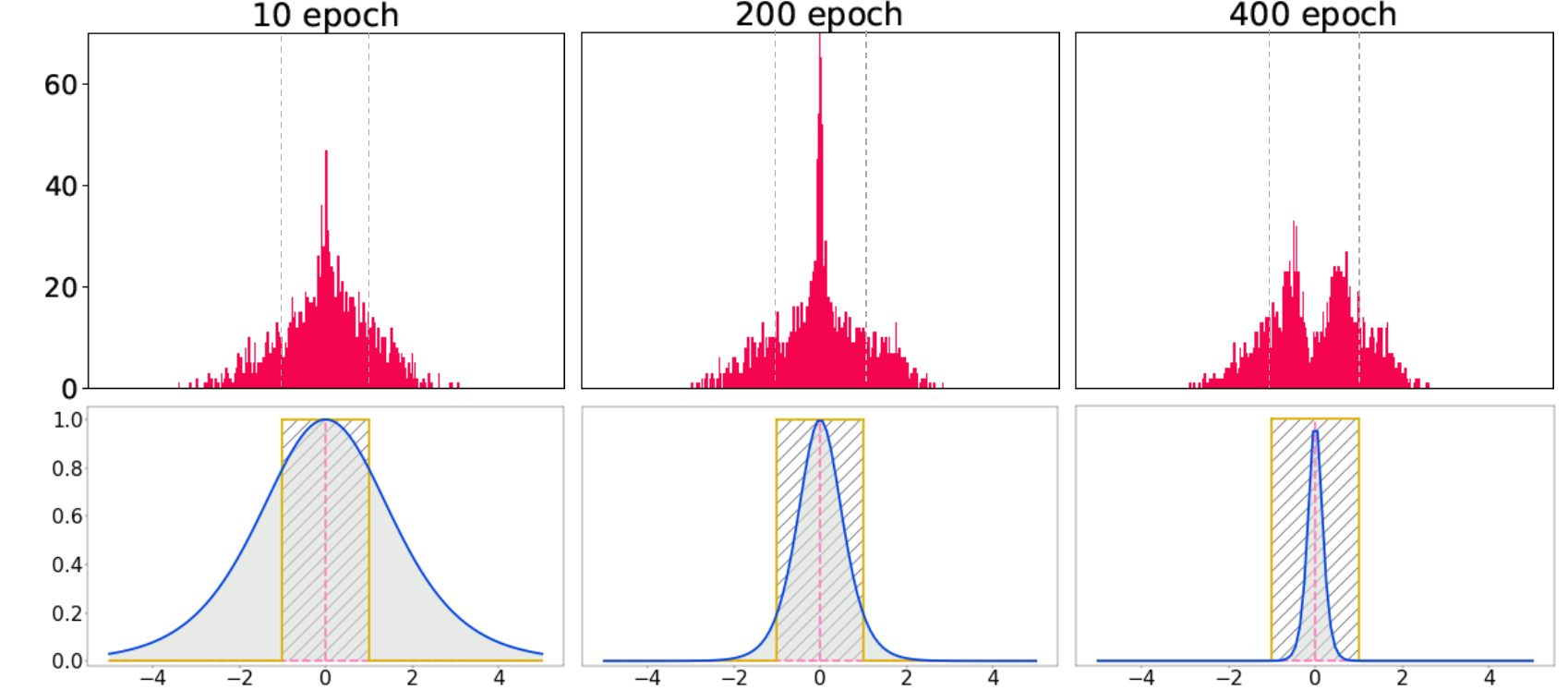}
	\end{center}
	\caption{EDE takes full account of the impact of weight distributions (after standardization) in different epochs (10, 200 and 400) during training. The weight distributions are shown in the upper part. In the lower part, the blue line is the derivative of EDE, the yellow and the pink lines are the derivative of STE and $\mathtt{sign}$ function, respectively. And the shade indicates the error between the derivative of EDE/STE and that of the $\mathtt{sign}$ function. In the first stage of EDE, all weights can be effectively updated. In the second stage, the weights around 0 are more accurately updated due to the decrease of gradient error.}
	\label{fig:ede_w_d}
	\vspace{-0.2in}
\end{figure}

To demonstrate the necessity and effect of our well-designed EDE, we show the data distribution of weights in different stages of training, as shown in Fig~\ref{fig:ede_w_d}. The figures in the first row show the distribution and the figures in the second row show the corresponding derivative curve. Among the derivative curves, the blue one represents EDE and the yellow one represents the common STE (with clipping). It can be seen that during the first stage of EDE (epoch 10 to epoch 200 in Fig~\ref{fig:ede_w_d}), there is much data outside the range [-1, +1], thus there should not be much clipping which will be harmful to update ability. Besides, the peakedness of weight distribution is high and much data gather around zero at the beginning of training. EDE keeps the derivative similar to $\mathtt{Identity}$ function at this stage to ensure the derivative around zero not too large, and thus avoids severely unstable training. Fortunately, with the binarization introduced into training, the weights will gradually approach -1/+1 in the later stages of training. Thus we can slowly increase the value of derivative and approximate a standard $\mathtt{sign}$ function to reduce the gradient mismatch.
The visualized results prove that our EDE approximation for backward propagation agrees with the real data distribution, which is the key to improving accuracy. 

\subsubsection{Ablation Performance}

We further investigate the performance using different parts of IR-Net with the ResNet-20 model on CIFAR-10, which helps understand how our IR-Net works in practice. Table~\ref{ablation_exp} shows the performance in different settings. From the table, we can see that using Libra-PB or EDE alone can improve the accuracy, and the weight standardization in Libra-PB also plays an important role. Moreover, the improvements brought by these parts together can be superimposed, that is why our method can train highly accurate binarized models.

\begin{table}[htb]
    \caption{Ablation study for IR-Net.}
    \vspace{-0.14in}
    \label{ablation_exp}
	\centering
    \setlength{\tabcolsep}{1.2mm}{\small
    \begin{tabular}{lcc}
		\toprule
		Method&\tabincell{c}{Bit-width\\(W/A)}&Acc.(\%)\\
		\midrule
		FP&32/32&90.8\\
		\midrule
		Binary&1/1&83.8\\
		Libra-PB (without weight standardization)&1/1&84.3\\
		Libra-PB (without bit-shift scales)&1/1&84.6\\
		Libra-PB&1/1&84.9\\
		EDE&1/1&85.2\\
		{IR-Net} (Libra-PB \& EDE)&{1/1}&\textbf{86.5}\\
		\bottomrule
	\end{tabular}
	}
	\vspace{-0.2in}
\end{table}

\subsection{Comparison with SOTA methods}

We further comprehensively evaluate IR-Net by comparing it with the existing SOTA methods.

\subsubsection{CIFAR-10} Table~\ref{cifar} lists the performance using different methods on CIFAR-10, including RAD~\cite{Regularize-act-distribution} over ResNet-18 (based on~\cite{ResNet-18-project}), DoReFa-Net~\cite{dorefa}, LQ-Net~\cite{LQ-Net}, DSQ~\cite{DSQ} over ResNet-20 (based on~\cite{ResNet-20-project}) and BNN~\cite{hubara2016binarized}, LAB~\cite{Loss-Aware-BNN}, RAD, XNOR-Net~\cite{DBLP:conf/eccv/RastegariORF16} over VGG-Small. 
In all cases, our IR-Net obtains the best performance. More importantly, our method gets a significant improvement over the SOTA methods when using 1-bit weights and 1-bit activations (1W/1A), whether we use the original ResNet structure or the Bi-Real structure. For example, in the 1W/1A bit-width setting, compared with SOTA on ResNet-20, the absolute accuracy increase is as high as 2.4\%, and the gap to its full-precision (FP) counterpart is reduced to 4.3\%. 

\subsubsection{ImageNet} For the large-scale ImageNet dataset, we study the performance of IR-Net over ResNet-18 and ResNet-34. Table~\ref{imagenet} shows a number of SOTA quantization methods over ResNet-18 and ResNet-34, including BWN~\cite{DBLP:conf/eccv/RastegariORF16}, HWGQ~\cite{DBLP:journals/corr/abs-1708-08687}, TWN~\cite{DBLP:journals/corr/LiL16}, LQ-Net~\cite{LQ-Net}, DoReFa-Net~\cite{dorefa}, ABC-Net~\cite{ABCNet}, Bi-Real~\cite{Liu_2018_ECCV},
XNOR++~\cite{XNOR++}, BWHN~\cite{DBLP:journals/corr/abs-1802-02733}, SQ-BWN and SQ-TWN~\cite{Dong2017Learning}.
We can observe that when only quantizing weights over ResNet-18, IR-Net using 1-bit outperforms most other methods by large margins, and even surpasses TWN using 2-bit weights. And in the 1W/1A setting, the Top-1 accuracy of IR-Net is also significantly better than that of the SOTA methods (\eg, 58.1\% vs. 56.4\% for ResNet-18). The experimental results prove that our IR-Net is more competitive than the existed methods.

\begin{table}[tp!]
    \caption{Performance comparison with SOTA methods on CIFAR-10.}
    \vspace{-0.14in}
    \label{cifar}
	\centering
    \setlength{\tabcolsep}{3.3mm}{\small
    \begin{tabular}{llcc}
		\toprule
		Topology&Method&\tabincell{c}{Bit-width (W/A)}&Acc.(\%)\\
		\midrule
		\multirow{3}{*}{ResNet-18}&FP&32/32&93.0\\
		&RAD&1/1&90.5\\
		&Ours\footnotemark[1]&1/1&\textbf{91.5}\\
		\midrule
		\multirow{10}{*}{ResNet-20}&FP&32/32&91.7\\
		&DoReFa&1/1&79.3\\
		&DSQ&1/1&84.1\\
		&{Ours\footnotemark[1]}&{1/1}&\textbf{85.4}\\
		&{Ours\footnotemark[2]}&{1/1}&\textbf{86.5}\\
		\cmidrule(r){2-4}
		&FP&32/32&91.7\\
		&DoReFa&1/32&90.0\\
		&LQ-Net&1/32&90.1\\
		&DSQ&1/32&90.2\\
		&{Ours\footnotemark[1]}&{1/32}&\textbf{90.8}\\
		\midrule
		\multirow{6}{*}{VGG-Small}&FP&32/32&91.7\\
		&LAB&1/1&87.7\\
		&XNOR&1/1&89.8\\
		&BNN&1/1&89.9\\
		&RAD&1/1&90.0\\
		&{Ours}&{1/1}&\textbf{90.4}\\
		\bottomrule
	\end{tabular}}
	\vspace{-0.2in}
\end{table}

\subsection{Deployment Efficiency}
\label{deploy_efficiency}
To further validate the efficiency of IR-Net when deployed into the real-world mobile devices, we further implement our IR-Net on Raspberry Pi 3B, which has a 1.2 GHz 64-bit quad-core ARM Cortex-A53, and test its real speed in practice. We utilize the SIMD instruction SSHL on ARM NEON to make inference framework daBNN~\cite{zhang2019dabnn} compatible with our IR-Net. We have to point out that till now there have been very few studies that reported their inference speed in real-world devices, especially when using 1-bit binarization. In Table~\ref{table:eff}, we compare our IR-Net with the existing high-performance inference implementation including NCNN~\cite{ncnn} and DSQ~\cite{DSQ}. From the table, we can easily find that the inference speed of IR-Net is much faster, the model size of IR-Net can be greatly reduced and the bit-shift scales in IR-Net bring almost no extra inference time and storage consumption.

\begin{table}[tp!]
    \caption{Performance comparison with SOTA methods on ImageNet.}
    \label{imagenet}
    \vspace{-0.14in}
	\centering
	\begin{threeparttable}
	{\small
	\setlength{\tabcolsep}{1.2mm}{
	\begin{tabular}{llccc}
		\toprule
		Topology&Method&\tabincell{c}{Bit-width (W/A)}&Top-1(\%)&Top-5(\%)\\
		\midrule
		\multirow{14}{*}{ResNet-18}&FP&32/32&69.6&89.2\\
		&ABC-Net&1/1&42.7&67.6\\
		&XNOR&1/1&51.2&73.2\\
		&BNN+&1/1&53.0&72.6\\
		&DoReFa&1/2&53.4&--\\
		&Bi-Real&1/1&56.4&79.5\\
		&XNOR++&1/1&57.1&79.9\\
		&{Ours\footnotemark[2]}&{1/1}&\textbf{58.1}&\textbf{80.0}\\
		\cmidrule(r){2-5}
		&FP&32/32&69.6&89.2\\
		&SQ-BWN&1/32&58.4&81.6\\
		&BWN&1/32&60.8&83.0\\
		&HWGQ&1/32&61.3&83.2\\
		&TWN&2/32&61.8&84.2\\
		&SQ-TWN&2/32&63.8&85.7\\
		&BWHN&1/32&64.3&85.9\\
		&{Ours\footnotemark[1]}&{1/32}&\textbf{66.5}&\textbf{86.8}\\
		\midrule
		\multirow{6}{*}{ResNet-34}&FP&32/32&73.3&91.3\\
		&ABC-Net&1/1&52.4&76.5\\
		&Bi-Real&1/1&62.2&83.9\\
		&{Ours\footnotemark[2]}&{1/1}&\textbf{62.9}&\textbf{84.1}\\
		\cmidrule(r){2-5}
		&FP&32/32&73.3&91.3\\
		&{Ours\footnotemark[1]}&{1/32}&\textbf{70.4}&\textbf{89.5}\\
		\bottomrule
	\end{tabular}}
	}
	\end{threeparttable}
	\vspace{-0.18in}
\end{table}

\footnotetext[1]{Results of ResNet with normal structure~\cite{he2016deep}.}
\footnotetext[2]{Results of ResNet with Bi-Real structure~\cite{Liu_2018_ECCV}.}

\begin{table}[htb]
    \caption{Comparison of time cost of ResNet-18 with different bits (single thread).}
    \label{table:eff}
    \vspace{-0.14in}
	\centering
	\setlength{\tabcolsep}{0.9mm}\small{
	\begin{tabular}{lccc}
		\toprule
		Method&\tabincell{c}{Bit-width\\(W/A)}&Size (Mb)&Time (ms)\\
		\midrule
		FP&32/32& 46.77 & 1418.94 \\
		NCNN&8/8& -- & 935.51 \\
		DSQ&2/2& -- & 551.22 \\
		Ours (without bit-shift scales)&1/1&\textbf{4.20}&\textbf{252.16}\\
		Ours&1/1&\textbf{4.21}&\textbf{261.98}\\
		\bottomrule
	\end{tabular}}
	\vspace{-0.2in}
\end{table}

\section{Conclusion}
In this paper, we propose IR-Net to retain the information propagated in binary neural networks, mainly consisting of two novel techniques: Libra-PB for keeping diversity in forward propagation and EDE for reducing the gradient error in backward propagation. Libra-PB conducts a simple yet effective transformation on weights from the view of information entropy, which simultaneously reduces the information loss of both weights and activations, without additional operation on activations. Thus, the diversity of binary neural networks can be kept as much as possible and meanwhile the efficiency will not be harmed. Besides, the well-designed gradient estimator EDE retains the gradient information during backward propagation. Owing to the sufficient updating ability and accurate gradients, the performance with EDE surpasses that with STE by a large margin. Extensive experiments prove that the IR-Net consistently outperforms the existed state-of-the-art binary neural networks.

\ 

\noindent\textbf{Acknowledgement}\quad This work was supported by National Natural Science Foundation of China (61872021, 61690202), and Beijing Nova Program of Science and Technology (Z191100001119050).

\newpage

{\small
\bibliographystyle{ieee}
\bibliography{egbib}
}

\end{document}